\documentclass[preprint]{vgtc}               

\graphicspath{{figures/}{pictures/}{images/}{./}} 

\usepackage{times}                     

\usepackage{tabu}                      
\usepackage{booktabs}                  
\usepackage{lipsum}                    
\usepackage{mwe}                       

\usepackage{mathptmx}                  

\usepackage{amsfonts}
\usepackage{amsmath}
\usepackage{enumitem}
\usepackage{microtype}

\newcommand{\subhead}[1]{
  \noindent\textbf{#1}%
}

\newcommand{\compresslist}{
  \setlength{\itemsep}{0pt}
  \setlength{\parskip}{0pt}
  \setlength{\parsep}{0pt}	
}

\newcommand{\paperTitle}{DE-VAE: Revealing Uncertainty in Parametric and Inverse Projections\\with Variational Autoencoders using Differential Entropy}

\DeclareMathAlphabet{\altmathcal}{OMS}{cmsy}{m}{n}

\newcommand{\verticalLabel}[1]{\rotatebox{90}{\parbox{4.0cm}{\centering\textbf{#1}}}}

\onlineid{1023}

\vgtccategory{Research}

\vgtcinsertpkg

\ieeedoi{10.1109/UncertaintyVisualization68947.2025.00009}

\title{\paperTitle}

\author{Frederik L. Dennig\thanks{e-mail: frederik.dennig@uni-konstanz.de}\\ %
        \scriptsize University of Konstanz %
\and Daniel A. Keim\thanks{e-mail: keim@uni-konstanz.de}\\ %
     \scriptsize University of Konstanz %
}

\teaser{
  \centering
  \includegraphics[width=0.88\linewidth]{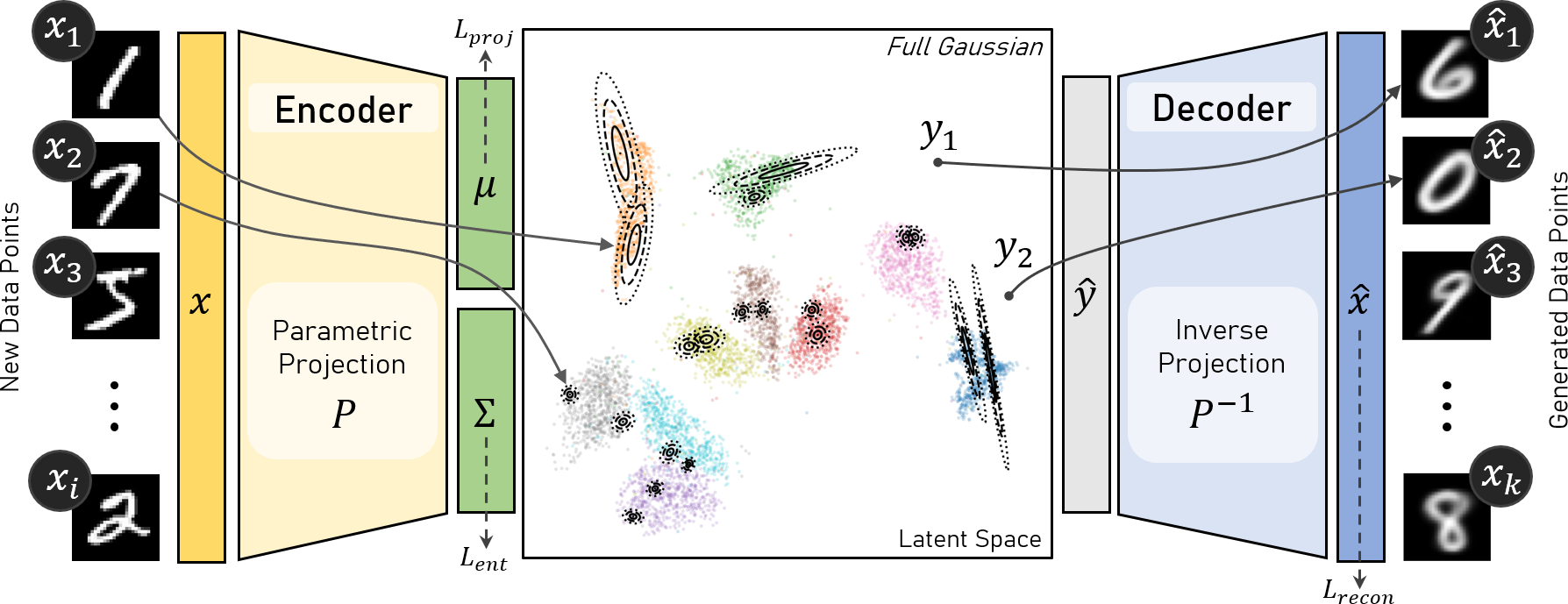}
  \caption{The encoder of a DE-VAE learns a parametric projection $P$ of MNIST, mapping each data point $x_i$ to a full Gaussian $\altmathcal{N}(\mu, \Sigma)$, modeling the uncertainty of a UMAP projection. The decoder learns an inverse projection $P^{-1}$, taking $y_k$ and reconstructing a plausible sample $\hat{x}_k$. $P$ enables uncertainty-aware visualization of the latent space. DE-VAEs optimize the losses: $\altmathcal{L}_{\text{recon}}$, ensuring reconstruction; $\altmathcal{L}_{\text{proj}}$, aligning $\mu$ with points of the projection; and $\altmathcal{L}_{\text{ent}}$, maximizing the variance of $\Sigma$. To show learned Gaussian distributions, we depict the 1st, 2nd, and 3rd standard deviations as ellipses around two randomly sampled points per class.}
  \label{fig:teaser}
}

\abstract{%
Recently, autoencoders (AEs) have gained interest for creating parametric and invertible projections of multidimensional data. Parametric projections make it possible to embed new, unseen samples without recalculating the entire projection, while invertible projections allow the synthesis of new data instances. However, existing methods perform poorly when dealing with out-of-distribution samples in either the data or embedding space. Thus, we propose DE-VAE, an uncertainty-aware variational AE using differential entropy (DE) to improve the learned parametric and invertible projections. Given a fixed projection, we train DE-VAE to learn a mapping into 2D space and an inverse mapping back to the original space. We conduct quantitative and qualitative evaluations on four well-known datasets, using UMAP and t-SNE as baseline projection methods. Our findings show that DE-VAE can create parametric and inverse projections with comparable accuracy to other current AE-based approaches while enabling the analysis of embedding uncertainty.%
} 


\begin{document}


\firstsection{Introduction}

\maketitle


Multidimensional projections, also referred to as dimensionality reduction (DR) techniques, are widely used and well-established tools for the visual exploration of high-dimensional data \cite{Jolliffe1986}.
DR methods are usually used to create 2- or 3-dimensional representations while trying to preserve relationships present in the high-dimensional space, e.g., distances and neighborhoods \cite{Nonato2019}.
In this context, \textit{parametric} projection methods employ a trained model to create a mapping from high- to low-dimensional space, enabling the projection of previously unseen data points, real and synthetic, without the need to recompute the projection \cite{Sainburg2021}.
\textit{Inverse} projection methods provide a usually smooth mapping from the projection space to the high-dimensional space, enabling the generation of new data from any position in the projection space \cite{Wijk2003}, e.g., helpful in visualizing and exploring the decision zones of classifiers.

However, DR is inherently \textit{lossy and prone to distortions}, such as false neighborhoods, or rifts, especially, in the case of non-linear DR methods \cite{Nonato2019}.
To address this, researchers proposed methods to quantify and visualize distortions and gradients between projected points.
Recent approaches for parametric and inverse projections, which are usually created using neural networks (NNs) \cite{Espadoto2021Unprojection} or variational autoencoders (VAEs) \cite{Dennig2025Evaluating}, did not address or leverage the \textit{uncertainty} inherent in the process.
In this work, we propose DE-VAEs, which use \textit{differential entropy} to capture the uncertainty in the parametric projection by modeling it in its latent space.
Thus, unlike traditional VAEs, DE-VAEs \textit{model} the uncertainty associated with a sample with a multidimensional Gaussian distribution.
In contrast, for the inverse- or generative-case, we \textit{sample} from the modeled distribution to retrieve similar synthetic data items.
Since DR for data projection is inherently visual \cite{Ngo2022}, we can gain insight into the sources and effects of uncertainty through uncertainty visualization \cite{Bonneau2014}.
Overall, we contribute the following:
\begin{enumerate}[label=(\arabic*),left=0pt]
\compresslist
\item We propose DE-VAEs, which leverage \textit{differential entropy} for creating parametric and invertible projections.
\item We perform an evaluation comparing \textit{isotropic}, \textit{diagonal} and \textit{full Gaussian} distributions quantitatively and qualitatively on four datasets using UMAP and t-SNE.
\item For \textit{reproducibility}, we provide the analysis, results and source code on \href{https://osf.io/zr6xf}{OSF} and \href{https://github.com/fredooo/DE-VAE}{GitHub}. 
\end{enumerate}
With this work, we aim to enhance high-dimensional data analysis by improving the interpretation of uncertainty in DR methods.

\section{Related Work}

First, we introduce some notation.
Let $D = \{x_{i}\}_{1 \leq i \leq n}$ denote a high-dimensional dataset with $d$ dimensions and $n$ samples, where each $x_i \in \mathbb{R}^d$.
A \emph{projection} technique $P$ transforms $D$ into a lower-dimensional set $P(D) = \{P(x_i) \mid x_i \in D\} = \{y_i\}_{1 \leq i \leq n}$, such that $P(D) \subset \mathbb{R}^q$ with $q \ll d$.
In our setting, $q = 2$, so $P(D)$ can be displayed as a 2D projection.

A plethora of DR methods exist and are studied in several surveys quantitatively and qualitatively \cite{Espadoto2019, Nonato2019}.
Generally, there are \textit{linear} and \textit{non-linear} methods which focus on either preserving \textit{global structure} or \textit{local neighborhoods}.
Linear projection techniques, such as principal component analysis (PCA) \cite{Jolliffe1986}, are computationally efficient and excel at maintaining the data’s global structure.
In contrast, non-linear global methods like multidimensional scaling (MDS) \cite{Kruskal1964, Leeuw2009} aim to capture more complex relationships while still focusing on the overall structure.
However, many modern non-linear approaches prioritize preserving local neighborhood relationships, often sacrificing global structure in the process.
Notable examples include t-SNE \cite{Maaten2008Tsne}, which forms implicit neighborhoods by computing pairwise similarities in high-dimensional space using a Gaussian kernel centered at each data point, while UMAP \cite{McInnes2018Umap} explicitly constructs a k-nearest neighbor graph to capture local structure and then optimizes a low-dimensional embedding that preserves the fuzzy topology implied by the graph.
AE-based models also belong to the class of non-linear methods; they learn low-dimensional embeddings by training neural networks to reconstruct input data, effectively capturing complex, non-linear feature representations \cite{Wang2016}.

\subhead{Parametric Projection Methods:} These approaches provide an explicit mapping function from the high-dimensional space to a lower-dimensional embedding, enabling efficient out-of-sample projections for new data points \cite{Hinton2006}.
Van der Maaten \cite{Maaten2009} employed a feed-forward neural network to create a parametric version of t-SNE. In a similar vein, parametric UMAP \cite{Sainburg2021}
utilizes neural networks, including autoencoders, as a parametric alternative to the original non-parametric embedding.
By training neural networks to predict 2D coordinates for input data points, Espadoto et al. \cite{Espadoto2020Deep} demonstrated that sufficiently large networks can effectively approximate various non-parametric methods.

\subhead{Inverse Projections:} A function $P^{-1}: \mathbb{R}^q \rightarrow \mathbb{R}^d$, that is typically smooth and aims to minimize the error $\sum_{x \in D} \|P^{-1}(P(x)) - x\|^2$, where $P$ is a projection and $\| \cdot \|$ denotes the $L_2$ norm, is called an inverse projection.
There are only a limited number of non-NN methods supporting inversion: PCA \cite{Jolliffe1986}, LAMP \cite{Joia2011, Amorim2012}, and UMAP \cite{McInnes2018Umap}. At the same time, an autoencoder, by design, has an encoder (learning $P$) and a decoder (learning $P^{-1}$) that reconstructs the original data~\cite{Bank2023}.
Other methods augment a subset of DR methods with inverse transformations, making different assumptions.
One of the first techniques used a global interpolation-based method to create synthetic data points \cite{Wijk2003}.
Amorim et al. \cite{Amorim2015} refined this approach by using a radial basis function for kernel-based interpolation.
Blumberg et al.~\cite{Blumberg2025MultiInv} invert MDS by applying data point multilateration based on geometrical properties.

\subhead{Uncertainty-aware Projection Methods}: These techniques support probability distributions modeling uncertainty as their input.
Görtler et al.~\cite{Goertler2020} proposed a linear DR technique that generalizes PCA to multivariate probability distributions by incorporating input uncertainty directly into the covariance matrix. 
UAMDS extends MDS by mapping high-dimensional random vectors from any distribution to a low-dimensional space supporting various stress types \cite{Haegele2023}.

\vspace{-0.25em}
\section{Differential Entropy in Variational Autoencoders}

In general, AEs are a type of NN for unsupervised learning of compressed data representations, i.e., latent representations \cite{Bank2023}, useful for feature extraction. 
In contrast, the latent representation of a variational autoencoder (VAE) is a probability distribution \cite{Kingma2014}.
Compared to AEs, VAEs learn a smooth and continuous latent space where nearby points correspond to similar outputs, useful for sampling and interpolation.
In recent work, VAEs were trained to jointly create a parametric projection $P$ and an inverse projection $P^{-1}$ \cite{Dennig2025Evaluating}.
The approach used a modified loss function for VAEs to structure the latent space, allowing for inverting a user-defined projection.
However, it did not outperform the more straightforward AE-based method.
We provide an alternative approach in this paper.
 
We show the general idea of a DE-VAE in \cref{fig:teaser}.
A DE-VAE learns to project data items from a ground-truth projection, i.e., a precomputed projection.
In contrast to existing parametric projection methods, DE-VAEs infer Gaussian distributions as their \textit{encoder} output, not just a low-dimensional coordinate.
The \textit{decoder} can reconstruct a high-dimensional data point from any 2-dimensional position.
More specifically, we target three objectives:
(1) Differing from VAEs, we target a latent mean $\mu$ alignment such that the \textit{encoder} learns a parametric projection $P$.
(2) We model uncertainty through variance via differential entropy.
(3) Similar to VAEs, we target accurate inverse projection $P^{-1}$ or reconstruction of the input data through the \textit{decoder}. 
Unlike the VAE's latent space, which is shaped by the Kullback-Leibler ($D_{KL}$) loss term, tying both mean and variance to a \textit{fixed prior} (e.g., $\altmathcal{N}(0, I)$), our approach independently controls the mean $\mu$ to align it with a projection $P(D)$. 
Additionally, the encoder infers an associated variance $\Sigma$ around the projected data point. 
In contrast to classical variational inference with respect to a prior, our regularization toward an externally chosen target shapes the latent space according to a projection.
In the following, we describe its loss function (see \cref{eq:overall-loss}) in detail.
Let $x \in D \subset \mathbb{R}^d$ denote the original vector, $\hat{x} \in \mathbb{R}^d$ the reconstructed vector output, $\mu \in \mathbb{R}^q$ the latent mean from encoder,
$y \in \mathbb{R}^q$ is the target latent vector, i.e. the embedding vector of the projection, and the Gaussian approximate posterior distribution $q(z) := q(z \mid x) = \altmathcal{N}(z \mid \mu(x), \Sigma(x))$.
\begin{align}\label{eq:overall-loss}
\begin{split}
\altmathcal{L}(x, \hat{x}, y, \mu, \Sigma) = \altmathcal{L}_{\text{recon}}&(x, \hat{x}) + \lambda_{\text{proj}} \cdot \altmathcal{L}_{\text{proj}}(y, \mu) \\
&+ \lambda_{\text{ent}} \cdot \altmathcal{L}_{\text{ent}}(\Sigma)
\end{split}
\end{align}
\subhead{Reconstruction Loss:} $\altmathcal{L}_{\text{recon}}(x, \hat{x})$ measures how well the model reconstructs the original input $x$ from its latent representation. Minimizing this term ensures that the latent code $z$ retains sufficient information to reproduce $x$ faithfully \cite{Kingma2014}. We refrain from a concrete choice since it is dataset-dependent, e.g., we use the mean squared error (MSE) and binary cross-entropy (BCE).

\subhead{Projection Loss:} $\altmathcal{L}_{\text{proj}}(y, \mu) = \| y - \mu \|^2$ encourages the latent mean $\mu$ to align with a target low-dimensional projection $y$, also called the ground-truth projection. This guides the latent space to reflect the structure of the projection $P(D)$. We use the MSE to measure the numerical difference between predicted and actual values.

\subhead{Entropy Term:} $\altmathcal{L}_{\text{ent}}(\Sigma) = -\altmathcal{H}[\altmathcal{N}(0,\Sigma)]$ promotes higher uncertainty or diversity in the associated latent distribution $\altmathcal{N}(\mu,\Sigma)$ by maximizing \textit{differential entropy}. This prevents overly confident or degenerate latent representations and supports smoother, more flexible embeddings. It needs to be carefully balanced, usually with $\altmathcal{L}_{rec}$, not to dominate the overall loss (see \cref{eq:overall-loss}).

\smallskip

Weighting terms are standard in loss functions of VAEs \cite{Higgins2017}.
We introduce $\lambda_{\text{proj}}$, which controls the strength with which the mean values of the latent space distributions will match the ground-truth projection, and $\lambda_{\text{ent}}$, which, with higher values increases the variance of each of the learned Gaussian distributions. $\lambda_{\text{ent}}$ is the more sensitive of the two and can overpower all other loss components.
We discuss the choice of $\lambda_{\text{proj}}$ and $\lambda_{\text{ent}}$ in \cref{sec:discussion}.

\vspace{-0.25em}
\subsection{Latent Gaussian Covariance Structures}

In the context of VAEs, $q(z)$ represents the \textit{approximate posterior distribution} over the latent variables $z$, given the observed data $x$, i.e., $ q(z) \equiv q_\phi(z \mid x) = \altmathcal{N}(z \mid \mu(x), \Sigma(x))$ with model parameters $\phi$. In the following, we briefly describe three different Gaussian distributions and how they impact the structure of the latent space (see supplementary material for mathematical deductions).

\begin{figure}[t]
    \setlength{\tabcolsep}{0.5mm}
    \begin{tabular}{ccc}
     & \textbf{MNIST (UMAP)} & \textbf{KMNIST (t-SNE)} \\
     \verticalLabel{Isotropic Gaussian} & \includegraphics[width=0.47\linewidth,trim=82 10 82 10,clip]{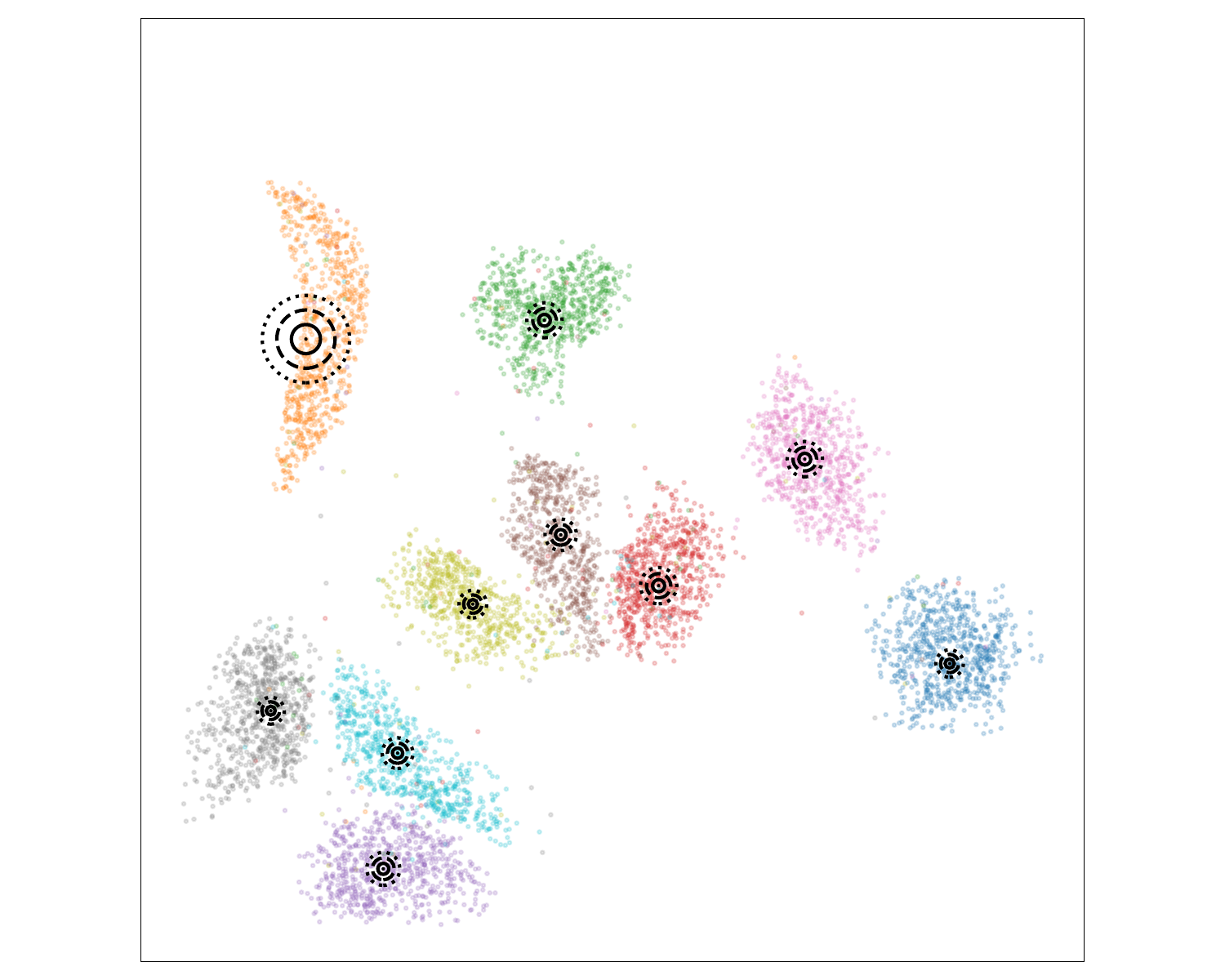} & \includegraphics[width=0.47\linewidth,trim=82 10 82 10,clip]{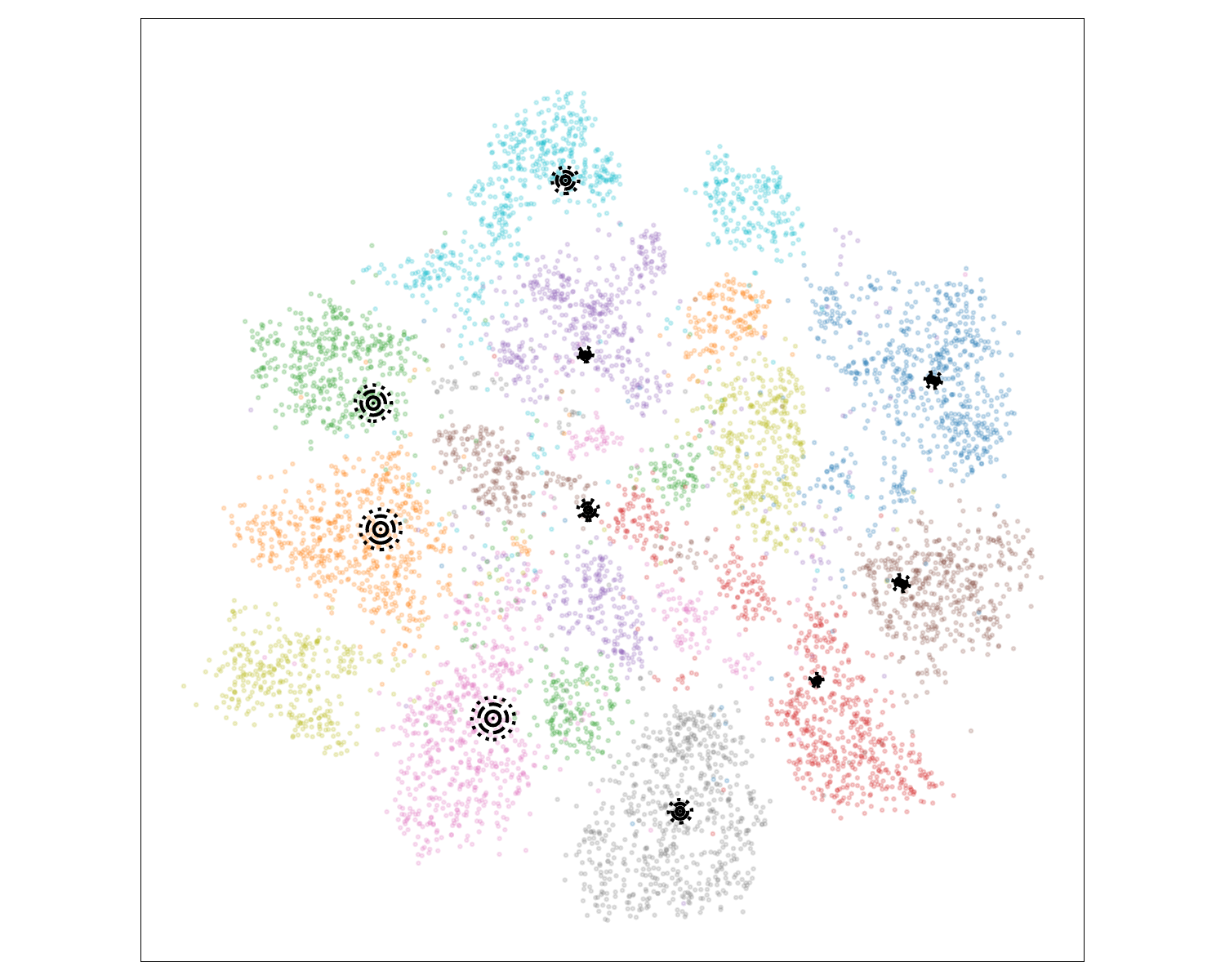} \\
     \verticalLabel{Diagonal Gaussian} & \includegraphics[width=0.47\linewidth,trim=82 10 82 10,clip]{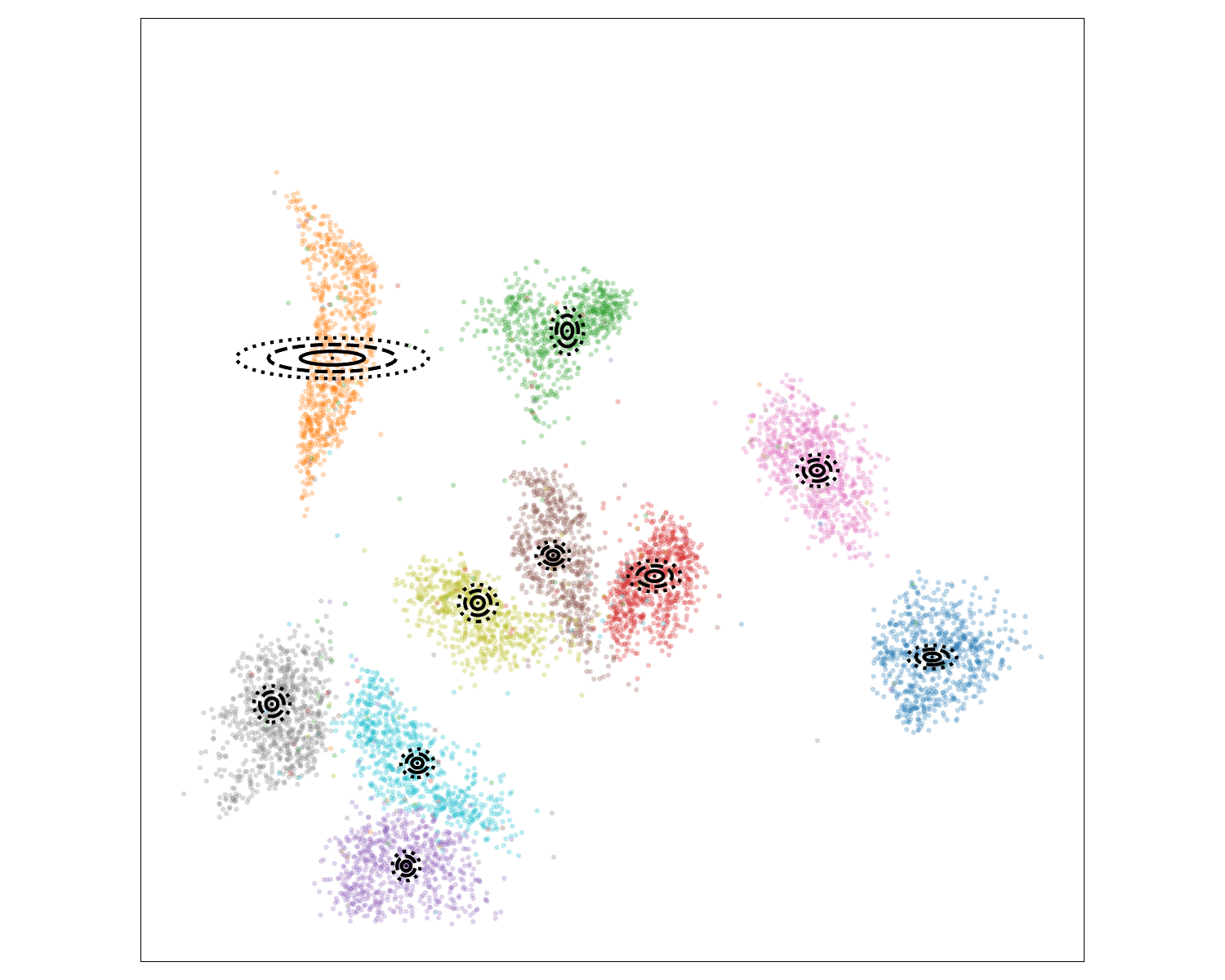} & \includegraphics[width=0.47\linewidth,trim=82 10 82 10,clip]{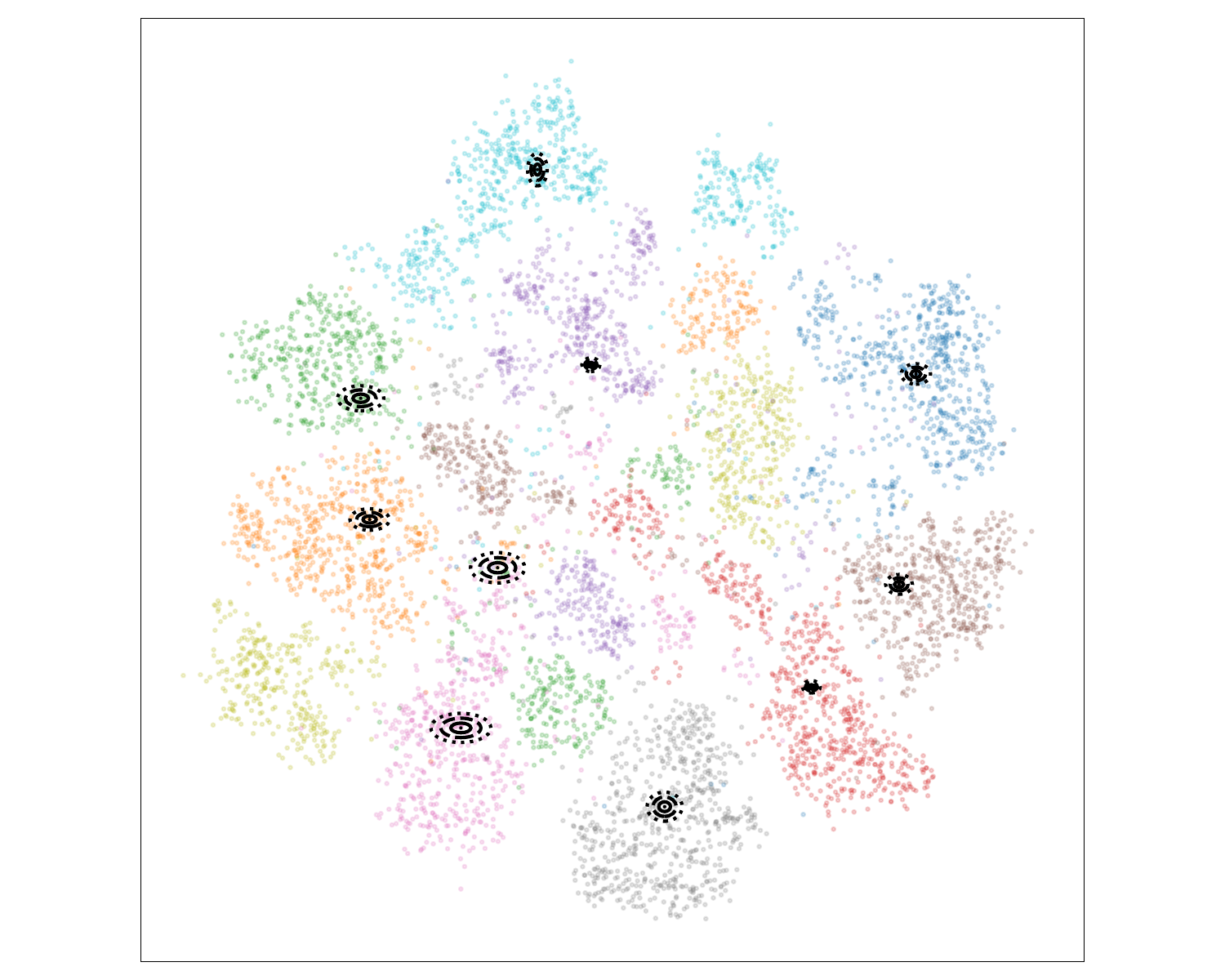} \\
     \verticalLabel{Full Gaussian} & \includegraphics[width=0.47\linewidth,trim=82 10 82 10,clip]{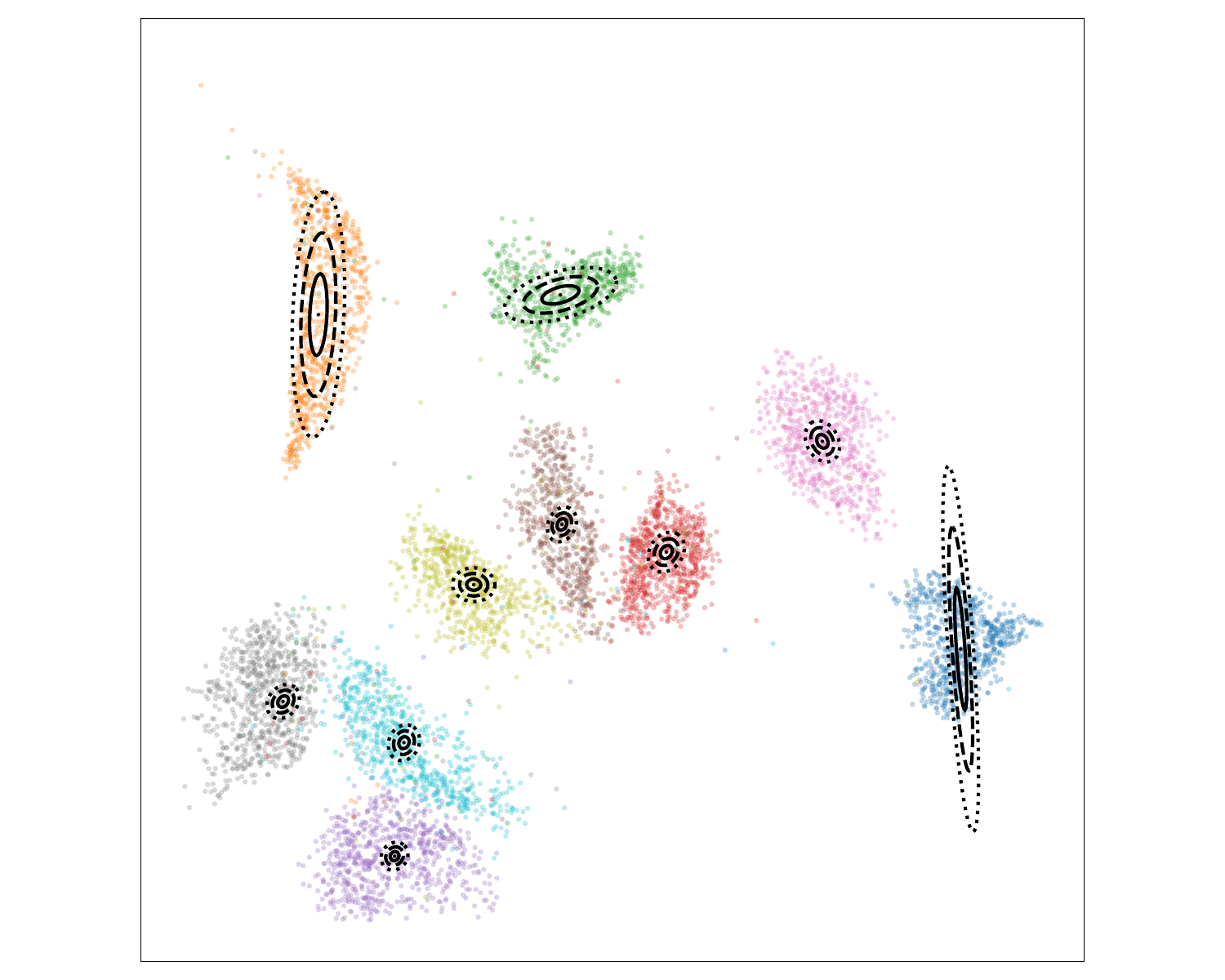} & \includegraphics[width=0.47\linewidth,trim=82 10 82 10,clip]{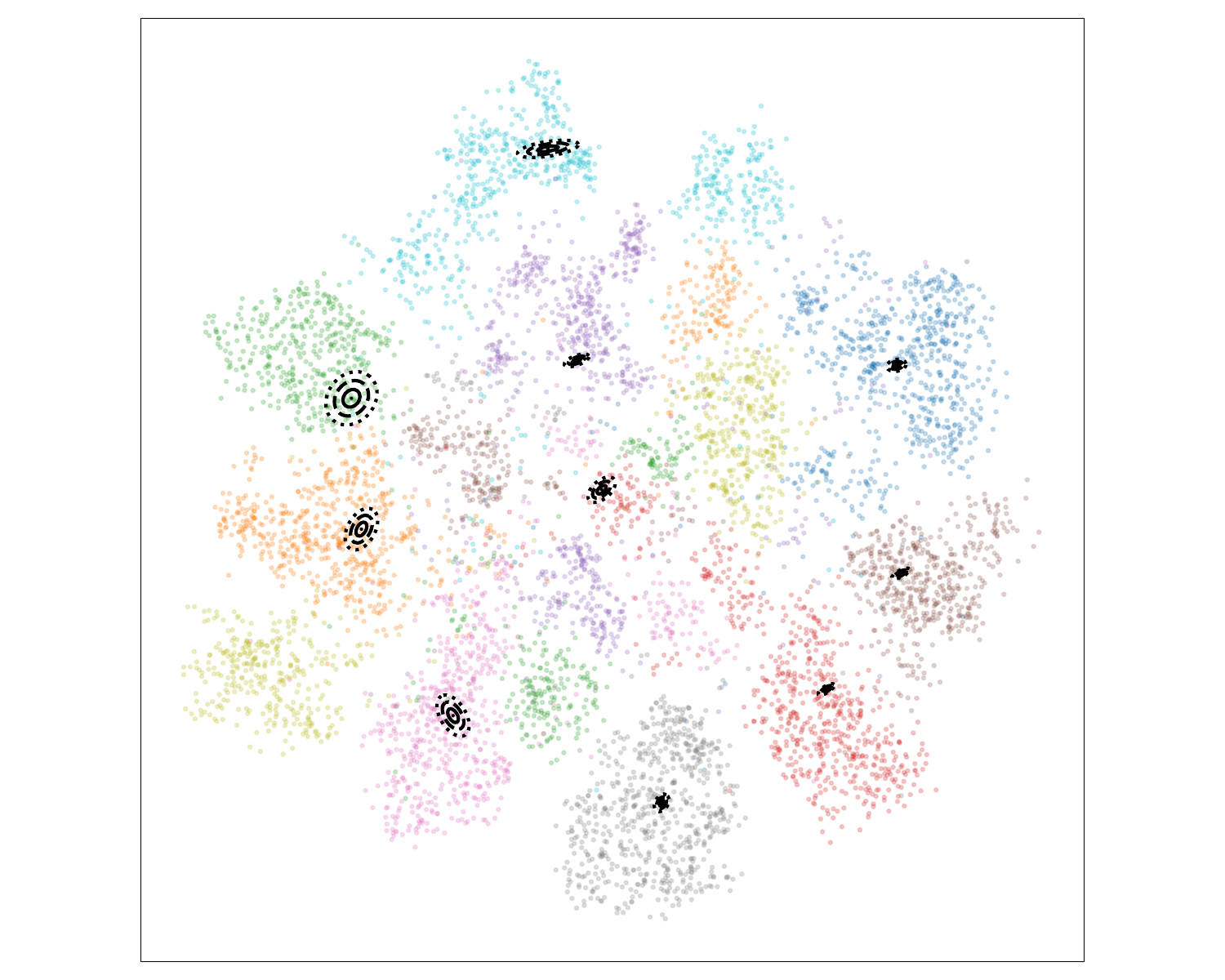}
    \end{tabular}
    \caption{We show the encoder output $\mu$ for all test data of the learned UMAP projection of MNIST and the t-SNE projection of KMNIST. The uncertainty modeling, expressed by three Gaussian distribution types, allows for varying levels of expressiveness through their differing degrees of freedom. These are shown as 1st, 2nd, and 3rd standard deviations, depicted as ellipses around the class medoids.}
    \label{fig:qualitative-results-latent-space}
    \vspace{-1.5em}
\end{figure}

\subhead{Isotropic Gaussian:}  
The covariance matrix assumes the same variance across all latent dimensions (see \cref{fig:qualitative-results-latent-space}), expressed as $\Sigma = \sigma^2 I$, where $I$ is the identity matrix. This simplification implies that latent variables are identically distributed with equal uncertainty. The differential entropy reduces to a form proportional to the logarithm of the shared variance, which directly contributes to the entropy term in the loss function. The entropy of an isotropic Gaussian $\altmathcal{N}(\mu, \sigma^2 I)$ is defined as $H[\altmathcal{N}(\mu, \sigma^2 I)] = \frac{q}{2} (1 + \log 2\pi + \log \sigma^2)$.

\subhead{Diagonal Gaussian:}  
In this case, each latent dimension has its own variance, so $\Sigma = \text{diag}(\sigma_1^2, \sigma_2^2, \dots, \sigma_q^2)$. This enables modeling of different uncertainty per dimension while maintaining independence between dimensions, i.e., axis-aligned (see \cref{fig:qualitative-results-latent-space}). The entropy term in the loss sums contributions from each dimension's variance, reflecting dimension-specific uncertainty in the latent space. The entropy of a diagonal Gaussian is defined as $\altmathcal{H}[\altmathcal{N}(\mu, diag(\sigma^2_1, \dots, \sigma^2_q))] = \frac{1}{2} \sum_i (1 + \log 2\pi + \log \sigma_i^2)$.

\subhead{Full Gaussian:}
This is the most general form with covariance $\Sigma = LL^\top$, where $L$ is a lower-triangular matrix obtained via the \textit{Cholesky decomposition}, which captures correlations between latent dimensions, allowing for more flexible uncertainty modeling (see \cref{fig:qualitative-results-latent-space}). The differential entropy is computed using the log-determinant of $\Sigma$, which in turn is derived from $L$, encouraging expressive latent representations through the covariance structure. The differential entropy of a full Gaussian is defined as $\altmathcal{H}[\altmathcal{N}(\mu, L)] = \frac{q}{2} (1 + \log 2\pi) + \sum \log L_{ii}$.

\vspace{-0.25em}
\section{Evaluation}

We compare our approach using four datasets, characterized by the number of samples $\mathbf{n}$, dimensions $\mathbf{d}$, intrinsic dimensionality $\mathbf{\rho_d}$, sparsity $\gamma_\mathbf{n}$, and \textbf{Type}.
We use \textit{HAR} \cite{Anguita2012} ($\mathbf{n}$: $10299$, $\mathbf{d}$: $561$, $\mathbf{\rho_d}$: $120$, $\gamma_\mathbf{n}$: $0.0\%$, \textbf{Type}: Sensor Data), \textit{MNIST} \cite{Lecun2010} ($\mathbf{n}$: $70000$, $\mathbf{d}$: $784$, $\mathbf{\rho_d}$: $330$, $\gamma_\mathbf{n}$: $82.9\%$, \textbf{Type}: Images), 
\textit{Fashion-MNIST} \cite{Xiao2017} ($\mathbf{n}$: $70000$, $\mathbf{d}$: $784$, $\mathbf{\rho_d}$: $187$, $\gamma_\mathbf{n}$: $50.2\%$, \textbf{Type}: Images), and
\textit{KMNIST} \cite{Clanuwat2018} ($\mathbf{n}$: $70000$, $\mathbf{d}$: $784$, $\mathbf{\rho_d}$: $238$, $\gamma_\mathbf{n}$: $66.7\%$, \textbf{Type}: Images).
We evaluate our approach quantitatively by comparing reconstruction loss and projection loss, as well as runtime measurements such as the number of training epochs and training time, using UMAP \cite{McInnes2018Umap} and t-SNE \cite{Maaten2008Tsne} with standard parameters as the ground-truth projections.
We also compare the approaches qualitatively by visually inspecting the latent space and reconstructions.
\textbf{None} refers to an AE without a variational component \cite{Dennig2025Evaluating}, predicting only a latent coordinate (equivalent to $\mu$), and serving as the current competitor in all evaluations.
Further details are provided in the supplementary material.
Replication data is available on \href{https://osf.io/zr6xf}{OSF} and \href{https://doi.org/10.18419/DARUS-5258}{DaRUS} \cite{Dennig2025DE-VAE-DaRUS}.

\begin{table}[t]
\setlength{\tabcolsep}{1.0mm}
\small
\centering
\begin{tabular}{lcccc}
& \textbf{None} & \textbf{Isotropic} & \textbf{Diagonal} & \textbf{Full} \\
\hline
\hline
\multicolumn{5}{c}{\textit{Parametric projection: Average projection loss $\altmathcal{L}_\text{proj}$ (lower is better)}} \\
\hline
\textbf{HAR (UMAP)} & $.675 \pm .685$ & $.525 \pm .257$ & $.447 \pm 0.166$ & $\mathbf{.404 \pm .079}$ \\
\textbf{MNIST (UMAP)} & $.402 \pm .064$ & $\mathbf{.400 \pm .077}$ & $.418 \pm .162$ & $.428 \pm 0.07$ \\
\textbf{Fashion-MNIST (UMAP)} & $\mathbf{.263 \pm .034}$ & $.293 \pm .069$ & $.306 \pm .059$ & $.298 \pm 0.05$ \\
\textbf{KMNIST (t-SNE)} & $\mathbf{30.4 \pm 4.68}$ & $32.3 \pm 3.98$ & $31.6 \pm 6.23$ & $33.0 \pm 5.27$ \\
\hline
\multicolumn{5}{c}{\textit{Inverse Projection: Average reconstruction loss $\altmathcal{L}_\text{recon}$ (lower is better)}} \\
\hline
\textbf{HAR (UMAP)} & $272 \pm 2.41$ & $272 \pm 2.42$ & $272 \pm 2.45$ & $273 \pm 4.73$ \\
\textbf{MNIST (UMAP)} & $ \mathbf{127 \pm 0.85}$ & $135 \pm 1.01$ & $136 \pm 1.64$ & $138 \pm 1.28$ \\
\textbf{Fashion-MNIST (UMAP)} & $\mathbf{246 \pm 1.93}$ & $252 \pm 2.36$ & $251 \pm 1.62$ & $253 \pm 1.53$ \\
\textbf{KMNIST (t-SNE)} & $\mathbf{239 \pm 1.64}$ & $246 \pm 1.83$ & $245 \pm 2.52$ & $248 \pm 3.02$ \\
\hline
\multicolumn{5}{c}{\textit{Number of training epochs until validation loss convergence (lower is better)}} \\
\hline
\textbf{HAR (UMAP)} & $41.3 \pm 13.1$ & $42.3 \pm 12.3$ & $41.2 \pm 10.9$ & $\mathbf{40.3 \pm 9.41}$ \\
\textbf{MNIST (UMAP)} & $32.8 \pm 6.54$ & $31.0 \pm 4.49$ & $32.8 \pm 6.54$ & $\mathbf{30.3 \pm 6.44}$ \\
\textbf{Fashion-MNIST (UMAP)} & $29.7 \pm 7.50$ & $29.0 \pm 11.0$ & $32.8 \pm 13.9$ & $\mathbf{28.4 \pm 8.46}$ \\
\textbf{KMNIST (t-SNE)} & $46.3 \pm 9.84$ & $\mathbf{38.5 \pm 5.75}$ & $45.9 \pm 11.4$ & $39.0 \pm 9.65$ \\
\end{tabular}
\vspace{0.5em}
\caption{Average losses and standard deviations (after $\pm$) of the parametric and inverse projections on test data for 10 runs each, as well as average number of training epochs to compare training time.
}
\label{tab:experiment-data}
\vspace*{-2.5em}
\end{table}

\vspace{-0.25em}
\subsection{Data Processing and Training}
 
We use the \emph{Adam optimizer} \cite{Kingma2015} with default parameters.
We set the learning rate to 0.001 and the batch size to 64, which are typical values.
We set the number of training epochs to a maximum of 100, allowing training to stop early if the validation loss does not improve for 5 consecutive epochs.
We determined the topologies of our AEs and the weighting factors of the loss function $\lambda_\text{proj}$ and $\lambda_\text{ent}$ through multiple runs, using the validation loss to track the impact of different parameter configurations.
For \textit{HAR}, we set $\lambda_\text{proj} = 5$ and $\lambda_\text{ent} = 0.001$.
For \textit{MNIST}, we set $\lambda_\text{proj} = 20$ and $\lambda_\text{ent} = 5$.
For \textit{Fashion-MNIST}, we set $\lambda_\text{proj} = 20$ and $\lambda_\text{ent} = 4$.
Finally, for \textit{KMNIST}, we set $\lambda_\text{proj} = 20$ and $\lambda_\text{ent} = 3$.
For \textit{HAR}, we used the vector data \textit{as-is} while using the MSE as the reconstruction loss.
For \textit{MNIST}, \textit{Fashion-MNIST}, and \textit{KMNIST}, we converted the pixel values into the interval $[0, 1]$, enabling us to use binary cross-entropy (BCE) as the reconstruction loss, $\altmathcal{L}_{recon}$.
In all cases, we used the projection methods with their default parameters without any additional post-processing.
For all details, refer to the source code.

\begin{figure*}[t]
    \centering
    \begin{minipage}[t]{0.185\linewidth}
    \centering
    \includegraphics[width=\linewidth]{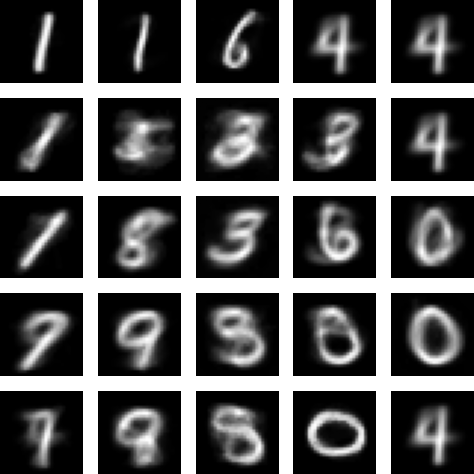} \\
    (a) UMAP (inverse)
    \end{minipage}
    \hfill
    \begin{minipage}[t]{0.185\linewidth}
    \centering
    \includegraphics[width=\linewidth]{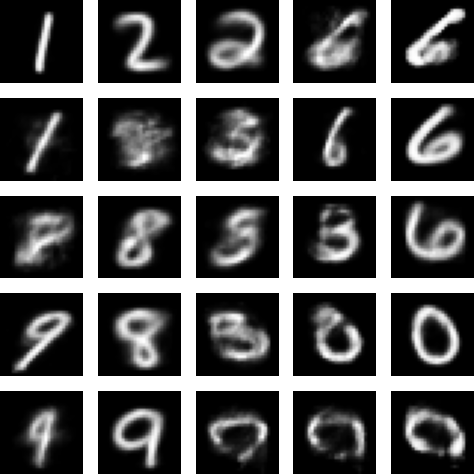} \\
    (b) None (AE)
    \end{minipage}
    \hfill
    \begin{minipage}[t]{0.185\linewidth}
    \centering
    \includegraphics[width=\linewidth]{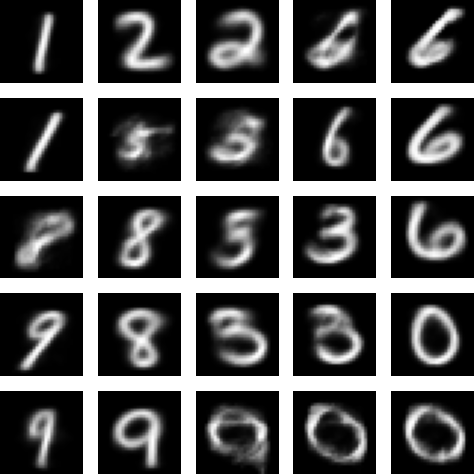} \\
    (c) Isotropic Gaussian
    \end{minipage}
    \hfill
    \begin{minipage}[t]{0.185\linewidth}
    \centering
    \includegraphics[width=\linewidth]{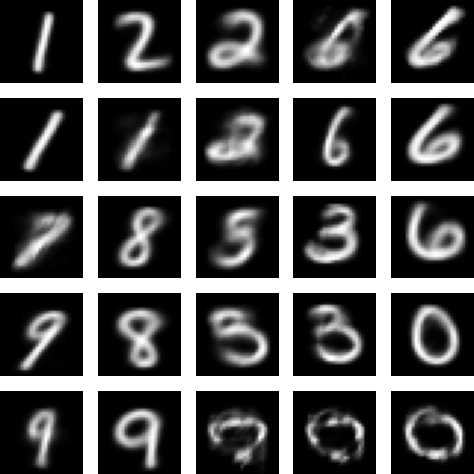}
    (d) Diagonal Gaussian
    \end{minipage}
    \hfill
    \begin{minipage}[t]{0.185\linewidth}
    \centering
    \includegraphics[width=\linewidth]{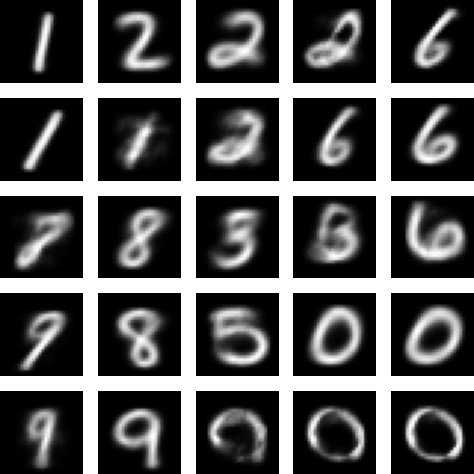}
    (e) Full Gaussian
    \end{minipage}
    \caption{For each approach (a-e), we inverse project 25 samples from an evenly spaced $5\times5$ grid on a UMAP projection of the \textit{MNIST} dataset and show the results of the inverse projections ($P^{-1}$). Our models (c-e) show higher quality and more diverse outputs.}
    \label{fig:qualitative-results-reconstruction}
    \vspace{-1em}
\end{figure*}

\vspace{-0.25em}
\subsection{Quantitative Comparison}

To evaluate the quality of the resulting parametric and inverse projections, we report $\altmathcal{L}_\text{proj}$ and $\altmathcal{L}_\text{recon}$ on the test samples of an 80/10/10 train-validation-test split for each method.
For \textit{MNIST}, \textit{Fashion-MNIST}, and \textit{KMNIST}, $\altmathcal{L}_\text{recon}$ is calculated using BCE, and MSE for \textit{HAR}.
In all cases, lower values indicate better performance.
$\altmathcal{L}_\text{proj}$ is always the MSE.
We train 10 AEs for each method using different random seeds for model initialization and batch randomization to mitigate outlying measurements. The average losses and the associated standard deviations are reported in \cref{tab:experiment-data}.

The average projection loss evaluates how accurately the model's latent space preserves structure from the projection. For \textit{HAR}, there is an improvement as the expressiveness of the covariance increases. The \textbf{Full} model achieves the best score, followed by \textbf{Diagonal}. The difference between \textbf{None} and the others is notable, suggesting that modeling latent uncertainty improves alignment with projections.
For \textit{MNIST}, the differences are very small. \textbf{Isotropic} slightly outperforms the others. However, all methods perform comparably and the standard deviations overlap.
For \textit{Fashion-MNIST} and \textit{KMNIST}, \textbf{None} performs best, with increasing projection loss for the other models, indicating over-regularization from uncertainty modeling.

The average reconstruction loss measures how well the decoder reconstructs inputs from latent samples. For \textit{HAR}, all methods perform identically, with negligible variation. For \textit{MNIST}, the plain AE performs best, with increasing loss from \textbf{Isotropic} to \textbf{Full}. This suggests that more expressive latent noise modeling slightly hurts reconstructions, possibly due to increased uncertainty. Also for \textit{Fashion-MNIST} and \textit{KMNIST}, \textbf{None} performs best, though the differences are small.
\textbf{None} or simple latent distributions are better for reconstruction.
This indicates that more complex models introduce uncertainty that can degrade decoder performance.

Comparing the number of training epochs, the \textbf{Full} models converge marginally faster, but the training time is longer overall.

\vspace*{-0.25em}
\subsection{Qualitative Comparison}

We visually compare the encoder output created by our three models (see \cref{fig:qualitative-results-latent-space}), depicting the 1st, 2nd, and 3rd standard deviations as ellipses around the class medoids, representing the learned Gaussians \cite{Haegele2023}.
As the quantitative analysis shows similar performance across models for the \textit{MNIST} and \textit{KMNIST} datasets for parametric projection, we can confirm that the methods do not differ significantly in the overall structure.
All clusters associated with each digit or character are visible.
We can also confirm that the choice of Gaussian distribution (i.e., isotropic, diagonal, or full) affects the learned distribution, as indicated by the rotated and strongly elliptical distributions, though less so for \textit{KMNIST}.

We also compare the reconstruction quality of the built-in inverse projection learned by the decoder (see \cref{fig:qualitative-results-reconstruction}).
We can observe that all approaches create clearer outputs than the inverse of UMAP (see \cref{fig:qualitative-results-reconstruction} (a)).
Our DE-VAEs (see \cref{fig:qualitative-results-reconstruction} (c)--(e)) also produce clearer outputs compared to the \textbf{None} (see \cref{fig:qualitative-results-reconstruction} (b)) that was recently proposed \cite{Dennig2025Evaluating}.
The comparison between DE-VAEs is not that clear. Given that the \textbf{Full} model is most expressive, via the full Gaussian, we can observe that the additional degrees of freedom do not impact reconstruction quality, at least for out-of-distribution samples.

\vspace*{-0.25em}
\section{Discussion and Limitations}\label{sec:discussion}

In principle, other spread measures exist beyond differential entropy. However, differential entropy provides an interpretable and differentiable measure of uncertainty that applies across different Gaussian families.
We acknowledge that an evaluation focusing on UMAP and t-SNE limits the generalizability of this work.
We used UMAP because it provides a built-in $P^{-1}$.
However, UMAP and t-SNE create strong cluster patterns with few to no outliers.

\subhead{Entropy-based Loss:}
While our loss shares similarities with that of a standard VAE, it does not fully satisfy the Evidence Lower Bound (ELBO). 
Our approach replaces the $D_{KL}$ with a differential entropy term, which encourages spread in the latent space but does not match a specific prior. Thus, our model can reconstruct inputs and preserve the projection, but does not support generative sampling from a prior. This may limit its ability to generate new data points from latent samples without additional regularization.

\subhead{Loss Weights $\lambda_{\text{proj}}$ and $\lambda_{\text{ent}}$:} We determined $\lambda_{\text{proj}}$ and $\lambda_{\text{ent}}$ experimentally by tracking the individual loss terms. We noticed that the other loss terms can be overpowered by a high $\lambda_{\text{ent}}$, which can cause the model to degenerate. Thus, it is key to balance $\lambda_{\text{ent}}$ with the reconstruction loss, such that $\altmathcal{L}_{\text{ent}}$ is still converging in the training process.
As a possible solution, we suggest annealing $\lambda_{\text{ent}}$ to maximize the entropy of the latent Gaussians. 
Generally, we found that models are less sensitive to high $\lambda_{\text{proj}}$.

\subhead{Future Work:}
We aim to broaden the evaluation to include additional projection techniques, such as MDS.  
We also intend to compare our AE-based models to established parametric and inverse projection methods.  
Using standard datasets and an 80/10/10 train-validation-test split, we plan to assess the robustness of each method with respect to training set size and intrinsic dimensionality.  
The differential entropy of the VAE serves as a form of implicit regularization for the latent space, which should yield a smooth inverse projection, but this effect has not yet been evaluated.  
The method can be extended to support layout enrichment \cite{Dennig2024} by combining the Gaussian distributions into a Gaussian mixture model.

\vspace*{-0.25em}
\section{Conclusion}

We proposed DE-VAEs, three VAEs with differential entropy-based structuring of the latent space with isotropic, diagonal, and full Gaussians, for generating \textit{parametric} and \textit{invertible} multidimensional data projections. The learned distributions differ notably across the models, thus providing a method to model uncertainty.
Qualitative evaluations with UMAP showed that DE-VAEs can outperform the other tested models regarding reconstruction quality for out-of-distribution samples.
Quantitatively, they did not consistently outperform competing approaches.
The parameterization of the loss function weights remains dependent on the dataset and projection.

\vspace*{-0.25em}
\acknowledgments{%
This work was funded by the Deutsche Forschungsgemeinschaft (DFG, German Research Foundation) -- Project-ID 251654672 -- TRR 161 (Project A03).%
}

\bibliographystyle{abbrv-doi-hyperref-narrow}

\bibliography{references}

\end{document}